%% file: Paper_R1.tex
\newcommand{\CLASSINPUTtoptextmargin}{22mm}
\newcommand{\CLASSINPUTbottomtextmargin}{18mm}
\newcommand{\CLASSINPUTinnersidemargin}{19mm}
\newcommand{\CLASSINPUToutersidemargin}{19mm}

\documentclass[letterpaper, 10 pt, conference]{IEEEtran}

\IEEEoverridecommandlockouts

\usepackage[dvipsnames, table]{xcolor}
\usepackage[T1]{fontenc}
\usepackage{amsmath,amssymb}
\usepackage{scalerel}
\usepackage{siunitx}
\usepackage{comment}
\usepackage{graphicx}
\usepackage{subfigure}
\usepackage{multirow}
\usepackage[ruled,vlined]{algorithm2e}
\usepackage{bm}
\usepackage{url}
\usepackage{hyperref}
\usepackage{diagbox} 
\usepackage{balance}
\usepackage{float}

\usepackage{eso-pic}
\newcommand\AtPageUpperMyright[1]{\AtPageUpperLeft{%
 \put(\LenToUnit{1.8cm},\LenToUnit{-1cm}){%
     \parbox{\textwidth}{\raggedleft\fontsize{9}{11}\selectfont #1}}%
 }}%
\newcommand{\conf}[1]{%
\AddToShipoutPictureBG*{%
\AtPageUpperMyright{#1}
}
}

\newtheorem{remark}{Remark}
\newcommand\changes[1]{\textcolor{.}{#1}}

\begin{document}

\title{\vspace{0.2cm}\LARGE \bf \changes{Exploring Deep Reinforcement Learning for Robust\\ Target Tracking using Micro Aerial Vehicles}}

\author{Alberto~Dionigi$^{1}$\thanks{$^{1}$ The authors are with the Department of Engineering, University of Perugia, 06125 Perugia, Italy {\tt\footnotesize \{alberto.dionigi, mirko.leomanni, gabriele.costante\}@unipg.it}.} \and Mirko~Leomanni$^{1}$ \and Alessandro~Saviolo$^{2}$\thanks{$^{2}$ The authors are with the Tandon School of Engineering, New York University, Brooklyn, NY 11201 USA {\tt\footnotesize \{alessandro.saviolo, loiannog\}@nyu.edu}.} \and Giuseppe~Loianno$^{2}$ \and Gabriele~Costante$^{1}$
\thanks{This work was supported by the NSF CAREER Award 2145277, the DARPA YFA Grant D22AP00156-00, the NSF CPS Grant CNS-2121391, Qualcomm Research, Nokia, and NYU Wireless.}
}

\maketitle
\conf{\centering This work has been accepted to the IEEE 21st International Conference on Advanced Robotics (ICAR). This is an archival version of our paper. Please cite the published version DOI: \url{https://doi.org/10.1109/ICAR58858.2023.10407017}.}

\begin{abstract}
\changes{The capability to autonomously track a non-cooperative target is a key technological requirement for micro aerial vehicles.}
In this paper, we propose an output feedback control scheme based on deep reinforcement learning for controlling a micro aerial vehicle to persistently track a \changes{flying target} while maintaining visual contact.
The proposed method leverages \changes{relative position data} for control, relaxing the assumption of having access to full state information which is typical of related approaches in literature.
Moreover, we exploit classical robustness indicators in the learning process through domain randomization to increase the robustness of the learned policy. 
Experimental results validate the proposed approach for target tracking, demonstrating \changes{high performance and robustness with respect to mass mismatches and control delays}. \changes{The resulting nonlinear controller significantly outperforms a standard model-based design in numerous off-nominal scenarios.}
\end{abstract}

\IEEEpeerreviewmaketitle

\changes{\section*{SUPPLEMENTARY MATERIAL}
\textbf{Video}: \url{https://youtu.be/22kI976fykA}}

\section{INTRODUCTION} \label{intro}
In recent years, Micro Aerial Vehicles (MAVs) like quadrotors have drawn significant attention for several applications including transportation, exploration, and surveillance due to their simplicity in design, agility, and low-cost~\cite{emran2018reviewquadrotor}.
A key feature of MAVs is their ability to hover in place and move in 3D which render them ideal platforms to persistently track flying targets.
The target tracking task requires a \textit{tracker} to follow a moving \textit{target} while maintaining a suitable attitude alignment (e.g., visual contact).
\changes{
This naturally leads to the formulation of an output feedback control problem, in which the relative position and the attitude motion are highly coupled. The resulting problem is hard to solve within a model-based control framework due to a number of factors.}
First, model-based methods require access to an accurate model of the MAV dynamics that is often hard to obtain.
This is a critical issue because MAVs are frequently affected by significant model uncertainties due to nonlinear effects generated by aerodynamic forces and torques, propeller interactions, payload variations, and communication delays~\cite{saviolo2022pitcn}.
Moreover, model-based control approaches usually rely on full pose information from an external motion capture system or an onboard estimator (see, e.g.,~\cite{mistler2001exact, casau2019robust, invernizzi2022global}). 
The former is only available in specific environments, while the latter is subject to non-negligible estimation errors that could significantly affect the performance~\cite{giurato2016quadrotor}. \changes{Finally, advanced techniques such as model predictive control usually require prior knowledge of the trajectory to be tracked~\cite{bicego2020nonlinear}, but this requirement is not met for the application at hand.
Due to these features and to the lack of systematic design techniques addressing the robust output feedback control problem for nonlinear systems, the applicability of model-based control to the considered target tracking task is currently limited to ad-hoc methods \cite{zhao2019robust}.}

\changes{To alleviate this limitation, one can employ a model-free control approach that directly leverages relative position measurement data collected by the tracker MAV, in a similar spirit to visual servoing~\cite{thomas2014ibvs}.}
In particular, an emerging paradigm deals with control algorithms based on Reinforcement Learning (RL)~\cite{sampedro2018image, polvara2018toward, xi2021anti}.
\changes{While robustness has been deeply analyzed in model-based control theory, only a few recent studies have investigated the possibility of learning robust controllers in a model-free fashion~\cite{molchanov2019sim, deshpande2021robust, turchetta2020robust}. In this paper, we exploit Deep Reinforcement Learning (DRL) to synthesize a MAV controller for robust target tracking.}

\input{figures/overview}
\subsection{Related Work}\label{related_work}
\changes{Proportional-Integral-Derivative (PID) control is by far the most commonly used control design for MAV applications, thanks to its ease of implementation. 
Many PID variants have been proposed in the literature, such as rotational and hierarchical controllers~\cite{Yu2015high,cao2015inner}. 
PID regulators provide adequate performance for simple set-point stabilization, but their application to more complex tasks, such as the one considered in this work, requires a specialized design and a careful selection of the tuning parameters. To overcome this issue, more advanced control techniques have been proposed in the literature. Based on the required level of exploitation of the vehicle dynamic model, these can be categorized into model-based and model-free.}

\changes{\textbf{Model-based control techniques.}
Model-based control is an attractive framework for MAV applications, as it allows one to rigorously find stability certificates.} Due to the highly nonlinear dynamics of MAVs, the majority of the contributions have focused on nonlinear techniques. \changes{ For example, \cite{mistler2001exact} employs a dynamic feedback linearization approach to solve the trajectory tracking problem via state feedback.} The main advantage of this method is that the design of the control law is carried out systematically, by exploiting results from linear systems theory. However, the resulting control policy can lack of robustness and this must be accounted for in the design process~\cite{mokhtari2005robust,lotufo2019control,leomanni2023robust}. Another viable solution is to adopt a backstepping design, so as to avoid the cancellation of useful nonlinearities ~\cite{cabecinhas2014nonlinear}. Sliding mode control is
a further design option providing enhanced robustness \cite{lee2009feedback}. More recently, \cite{casau2019robust,invernizzi2022global}  address the global stabilization problem by using hybrid state feedback control laws that extend the results in~\cite{lee2010geometric}.
These methods do not typically provide an easy way to optimize performance and to handle trajectory constraints.
To address this issue, \cite{saviolo2022pitcn,bicego2020nonlinear} propose a nonlinear model predictive control. However, this solution is computationally expensive and requires the reference trajectory to be available for prediction.

\changes{All the aforementioned works deal with trajectory tracking, i.e., with the problem of regulating the known MAV position towards a given time-varying reference. This problem is related to but simpler than the target tracking task considered herein. Indeed, systematic design techniques able to handle parametric uncertainty on the vehicle dynamics and stochastic measurement noise are not currently available for the nonlinear output feedback setting dictated by the considered task. This restricts the design options to ad-hoc control methods (see, e.g., \cite{zhao2019robust, wu2021vision}), for which robustness specifications are difficult to enforce at the synthesis level. One possibility to amend this drawback is to adopt a model-free framework.}

\changes{
\textbf{Model-free control techniques.} Model-free control techniques based on machine learning represent an emerging trend in robotics. In particular, RL approaches have been increasingly used in applications~\cite{li2020pose, devo2021enhancing, LEI2022}) and several recent works tackle the tracking problem for MAVs.}
\cite{hwangbo2017control} couples a multi-layer perceptron with a low-level PID controller to stabilize the MAV after starting from harsh initial conditions. 
\cite{lin2019supplementary} combines a classical feedback control law with a supplementary RL control policy. The control method is validated based on the vehicle's tracking performance.
While these works seek attitude stabilization through a standard controller, several approaches also consider learning end-to-end control using RL.
\cite{han2022cascade} proposes an end-to-end approach where a cascade control architecture based on RL is developed by linearizing the vehicle dynamics into six decoupled subsystems. While the linearization operation greatly simplifies the training process, it also significantly limits the maneuvering capability of the vehicle.
Other approaches address even more complex and dynamic tasks, such as aggressive flight \cite{sun2022aggressive} and decentralized swarm control \cite{batra2022decentralized}.
However, in these works, the input to the RL control policy includes highly privileged information, which may not be available in practice (e.g., absolute position) or need to be estimated separately (e.g., vehicle attitude). 
This limitation can be minimized by directly mapping on-board measurements into actuator commands~\cite{sampedro2018image, polvara2018toward, xi2021anti}.

\changes{
Existing RL-based approaches mainly focus on improving the tracking performance, and only some recent works start investigating the robustness of the learned policy to model uncertainties.}
\cite{molchanov2019sim,deshpande2021robust} propose robustification strategies based on domain randomization and max-min optimization, whereas
\cite{turchetta2020robust} employs the classical gain and delay margin indicators to improve the robustness of state feedback control design via RL.

\subsection{Contribution} \label{contribs}
A great deal of research work has been devoted to tracking problems involving MAVs. Model-free control via RL is still a relatively new topic and, in particular, the robustness issue has received less attention than the flight performance. Within this context,
this paper presents multiple contributions:
\begin{enumerate}
    \item \changes{We propose a systematic approach based on DRL to design a nonlinear output feedback control law for target tracking problems featuring visibility requirements}. The control law relies on relative position data that can be provided by onboard sensors, such as optical devices;
	\item The DRL policy is made robust against parametric uncertainties in the form of mass mismatches and time delays, by exploiting classical robustness indicators in the training process;
	\item We extensively evaluate the DRL policy in several tracking experiments \changes{and compare it to a baseline design employing feedback linearization and Linear-Quadratic-Gaussian (LQG) control}. The proposed approach outperforms consistently the baseline in off-nominal scenarios.
\end{enumerate}

\section{Preliminary Definitions} \label{sec:preliminary}
We leverage simulation to learn the control policy. This ensures unlimited training data, does not impose real-time constraints, and most importantly poses no physical risk to the robot.
In particular, we consider a surrogate model in which the tracker is controlled by thrust and angular velocity inputs. Following~\cite{mueller2015computationally}, the training model is defined as
\begin{equation}\label{sysmodel}
\begin{array}{c c l}
\ddot{p}(t)&=&  R_3(t)\dfrac{f(t)}{m} -g,\\[3mm]
\dot{R}(t)&=&R(t)\,[\omega(t)]_\times,
\end{array}
\end{equation}
where $p(t)$, $R(t)$ and $\omega(t)$ describe the tracker absolute position, orientation and angular velocity, respectively, $R_j(t)$ denotes the $j$-th column of $R(t)$, $f(t)$ is total thrust, $m$ is the vehicle mass, $g=[0\;0\;9.8]^\top \SI{}{\meter\per\second\squared}$ is the gravity vector, and $[\omega(t)]_\times$ is the skew-symmetric representation of $\omega(t)$. 
Thrust saturation constraints are included in the model by suitably clipping $f(t)$. 
\changes{We find it convenient to treat the thrust variation $\lambda(t)=\dot{f}(t)$ as a control input variable. 
Therefore, system \eqref{sysmodel} is augmented with the integrator
\begin{equation}\label{sysmodelf}
f(t)=f_0+\int_{t_0}^t\lambda(\tau) \;\text{d}\tau,
\end{equation}
where $f_0$ is the integration constant and $t_0=0$ is the initial time. The integral action provided by eq.~\eqref{sysmodelf} is exploited for controller design in order to reject constant disturbance accelerations (such as $g$).} 

Model uncertainty is taken into account by defining $m=\alpha m_0$ in eq.~\eqref{sysmodel}, where $m_0$ is the nominal mass and $\alpha$ is an uncertain gain parameter. Moreover, a time delay $\delta$ is added at the control level. This corresponds to specifying the inputs of the system eqs.~\eqref{sysmodel}-\eqref{sysmodelf} as follows
\begin{equation}\label{tdel}
\left[
\begin{array}{c}
\omega(t)\\
\lambda(t)
\end{array}
\right]=u(t-\delta)
\end{equation}
where $u(t)$ is the command provided by the controller. \changes{The choice of the control inputs in eq.~\eqref{tdel} is found to be effective for the design of high-performance controllers based on DRL (see, e.g., \cite{kaufmann2022bench}).}  The parameters $\alpha$ and $\delta$ will be used to robustify the learned control policy. 

The MAV dynamics are zero-order-hold discretized to obtain a discrete-time transition model suitable for RL. The sampling instants are denoted by $k=i t_s$, where $i$ is the discrete-time step number and $t_s$ is the sampling time. The motion of the target is modeled by a parameterized class of trajectories denoted by $p_r(k)$. These include fixed points and sampled sinusoidal signals with random amplitude, frequency, and phase.
The output of the model is specified as
\begin{equation}\label{sysout}
y(k)=R(k)^T [p_r(k)-p(k)]
\end{equation}
and describes the target position relative to the tracker, as seen from the tracker body-fixed frame. In the model-free framework discussed hereafter, the learning agent has only access to the output eq.~\eqref{sysout} and does not require information about the dynamic model presented in eq.~\eqref{sysmodel}. \changes{Notice that the output eq.~\eqref{sysout} can be measured by processing images from a depth or a stereo camera installed onboard the tracker with a suitable detection algorithm \cite{thomas2017autonomous}. The image detection process is not modeled in this paper. Instead, the measurements are generated by corrupting eq.~\eqref{sysout} with noise. This is a simplifying assumption that allows us to focus on the performance achievable by the controller. Moreover, it provides some generality, since it does not confine the analysis to a specific detector design.}

\section{Learning-based Control Approach}\label{approach}

\subsection{Problem Formulation}
The control objective for the {tracker} is to follow a moving target in such a way that visual contact is maintained. We aim at solving the target tracking problem with a controller driven by measurements of the output vector eq.~\eqref{sysout}. The controller must be robust to noise and model uncertainties. Moreover, it must avoid collisions between the tracker and the target while generating control commands compatible with the actuator saturation limits. 

A learning-based approach is adopted for controller design. More specifically, the controller consists of a DRL agent that interacts with the environment over a series of independent episodes and, based on the current observation $o(k)$, produces an action $u(k)$ and receives a reward $r(k)$. The observation $o(k)$ consists of the error between a sequence of noisy measurements of $y(k)$ and a predefined constant set-point $y_r$. \changes{The vector $y_r$ specifies the desired relative position between the target and the tracker, in the tracker body-fixed frame (e.g., a given location in the camera frame). Formally, let us define the tracking error}
\changes{\begin{equation} \label{eq:error_track}
e(k)=y_r-y(k),
\end{equation}
and the observation sequence
\begin{equation}
\label{eq:observation}
o(k)\!=\!\left[
\begin{array}{c}
e(k)+w(k)\\[1mm]
e(k-t_s)+w(k-t_s)\\[1mm]
\vdots \\[1mm]
e(k-H t_s)+w(k-Ht_s)
\end{array}\right],
\end{equation}}
\changes{where $H$ is the length of the sequence and $w(k) \sim \mathcal{N}(0,\, \sigma_w^2)$ is a Gaussian random noise. Then, the target tracking problem is cast as follows: steer $e(k)$ to zero, using the control law
\begin{equation}
\label{eq:output_policy}
u(k) = \pi(o(k)),
\end{equation}
where $\pi(\cdot)$ is a suitable control policy to be learned.
The control action $u(k)$ is assumed to take values in a continuous action space. The mapping between $u(k)$ and the actual MAV inputs is given by eqs.~\eqref{sysmodelf} and~\eqref{tdel}.
\begin{remark}
The considered target tracking task differs from standard MAV trajectory tracking. In particular, the latter requires regulating the known position vector $p(k)$ towards a given time-varying reference $p_r(k)$. {In the target tracking task considered, instead,} one has access only to the composite output $y(k)$, 
while the individual components on the right-hand side of eq.~\eqref{sysout}, as well as the future target positions $p_r(k+1), p_r(k+2), \dots$, are unknown. \changes{It is worth stressing that the relative position and the tracker attitude information are merged in the output eq.~\eqref{sysout}. While this makes the resulting output feedback control problem much more challenging, it can be exploited to regulate the tracker MAV position and attitude in a coordinated fashion, so as to maximize the visibility of the target. Indeed, this corresponds to driving $e(k)$ in eq.~\eqref{eq:error_track} towards zero.}
\end{remark}}

\subsection{\changes{Reward Shaping}}
The {reward} signal $r(k)$ is specifically designed to address the target tracking task and accounts for the salient features of the control problem. The main control objective is to steer the tracking error described by eq.~\eqref{eq:error_track}  to zero. To this purpose, the following contribution is defined
\begin{equation} \label{rew:track}
    r_{e}(k) = (r_x(k) \, r_y(k) \, r_z(k))^{\beta},
\end{equation}
where
\begin{equation} \label{rew:axis}
\begin{matrix}
r_x(k) = \max(0, 1 - \left | e_{1}(k) \right |), \\ 
r_y(k) = \max(0, 1 - \left | e_{2}(k) \right |), \\
r_z(k) = \max(0, 1 - \left | e_{3}(k) \right |),
\end{matrix}
\end{equation}
where $e_{j}(k)$ is the $j$-th entry of $e(k)$, and $\beta>0$ is a suitable exponent. The value of $r_{e}(k)$ is maximized when the tracking error is zero and it is clipped in the interval $\left [ 0, 1\right ]$ to favor the learning process. To ensure that the control effort is also optimized and the maximum velocity remains within reasonable limits while tracking, we define a velocity penalty $r_{v}(k)$ and a control effort penalty $r_{u}(k)$ as follows
\begin{equation} \label{pen:vel}
    r_{v}(k) = \frac{\| \dot{y}(k) \|}{1 + \| \dot{y}(k) \|},
\end{equation}
\begin{equation} \label{pen:u}
    r_{u}(k) = \frac{\| u(k) \|}{1 + \| u(k) \|}.
\end{equation}
Collision avoidance constraints are included by penalizing the RL agent with a very large negative reward whenever $\| y(k) \| < y_{m}$, where $y_{m}$ is the minimum distance allowed.

The {reward} function is obtained by adding up all the above contributions, which results in
\begin{equation} \label{rew:total}
r(k) = \left\{\begin{array}{l c}
r_{e}(k) \!-\! k_v  r_{v}(k)\! - \!k_u r_{u}(k) & \| y(k) \| > y_{m} \\[1mm] 
-c & \text{otherwise}
\end{array},\right.
\end{equation}
where $k_v>0$ and $k_u>0$ are weighting parameters that allow to trade-off between the reward terms, and $c$ is a large positive constant. Saturation limit constraints and robustness specifications are taken into account in the architecture design and learning phase as detailed next.

\subsection{Deep Reinforcement Learning Strategy} \label{sec:drl_strategy}
\changes{In the considered scenario, the tracking agent is free to operate in a three-dimensional space using continuous control actions. Due to the conspicuous dimension of the state space and the continuous nature of this problem, a classical tabular RL approach cannot be applied. Instead, we employ a DRL strategy that takes advantage of Deep Neural Network (DNN) approximators. In order to develop an effective target tracking policy  we adopt an \textit{asymmetric actor-critic} framework \cite{pinto2017asymmetric, dionigi2022vat}.
More specifically, according to this framework \cite{sutton2018reinforcement}, we design two DNN architectures: one for the \textit{actor} (A-DNN) and the other for the \textit{critic} (C-DNN). The former learns the optimal policy $\pi(o(k))$, while the latter is responsible for evaluating such a policy during training.} 

\changes{The A-DNN is a Multi-Layer Perceptron (MLP) with three hidden layers, each one composed of $256$ neurons and ReLU activations. The network input is a $3 H$-dimensional vector obtained by flattening $o(k)$ in eq.~\eqref{eq:observation}, while its output is the four-dimensional vector $u(k)$ representing the control commands of the tracker MAV.
The physical characteristics of the MAV motors impose saturation limit constraints on the control command. To enforce them on the A-DNN output, we add a $\tanh$ function to the last layer of the network. This keeps the action values computed by the {actor} in a fixed range (saturation limits are reported in Table \ref{table:hyper}).}

\changes{The C-DNN acts only in the training phase and, due to the asymmetric framework employed in this work, we are allowed to provide it with more privileged information with respect to that available to the A-DNN during inference. This training procedure favors the development of an effective criterion for policy evaluation. In particular, we define the observation $o_c(k)$ of the C-DNN as
follows
\begin{equation}
\label{eq:observation_critic}
o_c(k)\!=\!\left[
\begin{array}{c}
e(k)\\[1mm]
R(k)^T [\dot{p}(k)-\dot{p}_r(k)]\\[1mm]
R(k)^T [\ddot{p}(k)-\ddot{p}_r(k)]
\end{array}\right],
\end{equation}
to capture the instantaneous relative velocity and acceleration. The action $u(k)$ is also provided to the C-DNN in order to estimate the \textit{action-value} function $Q_\mathbf{\pi}(o_c(k), u(k))$. The C-DNN has the same structure as the A-DNN, except for two main differences: (i) the input is given by $o_c(k)$ in eq.~\eqref{eq:observation_critic} concatenated with $u(k)$ in eq.~\eqref{eq:output_policy}; (ii) the last layer of the network has a linear activation with the scalar output $Q_\mathbf{\pi}(o_c(k), u(k))$, i.e., the estimated {action-value}}.

\input{tables/hyper}

The A-DNN and C-DNN architectures are trained by using the popular Soft Actor-Critic (SAC) algorithm \cite{haarnoja2018soft}. In order to achieve robustness with respect to the model uncertainties, we employ domain randomization \cite{tobin2017domain} and initialize the parameters $\alpha$ and $\delta$ with random values at the beginning of each episode. Note that, in the context of linear systems theory, the latter parameters match the interpretation of the gain and phase margins. As shown in \cite{turchetta2020robust}, these indicators are still meaningful for nonlinear control design via RL. \changes{Therefore, it is expected that the learned policy exhibits robustness to such parametric uncertainties}. This is confirmed by the results in Section \ref{sec:ev_metrics}.
\section{EXPERIMENTS} \label{experiments}
In this section, we provide details about the training of the DRL agent and discuss the controller evaluation campaign. 

\subsection{\changes{Training Details}}
\label{sec:imlementation}

We trained the A-DNN and the C-DNN by using the Stable-Baselines3 \cite{stable-baselines3} implementation of SAC, which we customized to implement the asymmetric actor-critic algorithm\footnote{\url{https://github.com/isarlab-department-engineering/trackingMAV}}. 
The networks have been trained for a total of about $80,000$ episodes across $8$ parallel environments, by using the Adam optimizer with a learning rate of $0.0003$ and a batch size of $256$.
\input{figures/nominal_trajectory}
\input{figures/tracking_nominal}
\input{figures/perturbed_trajectory}
\input{figures/tracking_worstcase}%
\input{tables/results} 

The training session is structured in episodes and, at the beginning of each one, the tracker is placed inside the environment and starts from a hovering condition. \changes{During the episode, the target moves along a sinusoidal trajectory parameterized as follows
\begin{equation*}\label{referencetrajectory}
p_r(k) = p_r(0)+
\begin{bmatrix}
    A_x  \sin(2\pi f_x  k + \phi_x)\\ 
    A_y  \sin(2\pi f_y  k + \phi_y)\\ 
    A_z  \sin(2\pi f_z  k + \phi_z)    
\end{bmatrix}
-
\begin{bmatrix}
    A_x  \sin(\phi_x)\\ 
    A_y  \sin(\phi_y)\\ 
    A_z  \sin(\phi_z)    
\end{bmatrix},
\end{equation*}
where $p_r(0)$ is the target initial position and ($A_x$, $A_y$, $A_z$), ($f_x$, $f_y$, $f_z$), ($\phi_x$, $\phi_y$, $\phi_z$) are respectively the amplitude, the frequency, and the phase of the sinusoidal signals along the three axes. To produce a different trajectory for each episode, the trajectory parameters are randomized at the beginning and kept fixed for the entire duration of the episode. In particular, the target is spawned at the initial position $p_r(0)=p(0)+R(0)y_r+w_s$, where each entry of the random vector $w_s$ follows the Gaussian distribution $\mathcal{N}(0,\, \sigma_s^2)$. This promotes the development of tracking behaviors that are invariant to the initial condition. Furthermore, to achieve robustness to model uncertainties, we randomize the parameters $\alpha$ and $\delta$ by sampling them using a uniform distribution (see Table \ref{table:hyper} for a comprehensive list of the training hyper-parameters). The episode ends when one of the following conditions is met: (i) the step number $k$ reaches a predefined maximum limit; (ii) the collision constraint $\| y(k) \| > y_{m}$ is violated.} 

In the aforementioned setting, the optimization requires about 4 hours and 1.1GB of VRAM to converge on a workstation equipped with NVIDIA Quadro GV100 with 32GB of VRAM, an Intel Xeon Silver processor (2.40GHz ×24) and 128 GB of DDR4 RAM. At inference time, the VRAM required for the Actor Network is about 320kB, and the time required to compute the action is approximately $0.001$ s.

\subsection{Controller Validation} \label{sec:ev_metrics}

In order to evaluate the performance and robustness of the proposed DRL policy, we carried out an extensive simulation campaign featuring both nominal and off-nominal scenarios. To add more realism to the simulations, we defined a validation environment in which the system described by eq.~\eqref{sysmodel} is augmented by the angular velocity dynamics. \changes{These are stabilized by a low-level proportional controller that tracks the angular velocity command provided by the DRL agent. 
The resulting simulation model is given by
\begin{equation*}
\begin{bmatrix}
\ddot{p}(t)\\ 
\dot{R}(t)\\ 
\dot{\omega}_s(t)
\end{bmatrix} = 
\begin{bmatrix}
\frac{1}{m}(R_3(t)f(t)) - g\\ 
R(t)\left [  \omega_s(t)\right ]_{\times}\\ 
J^{-1}(k_{\omega}(\omega(t)-\omega_s(t))-\left [ \omega_s(t) \right ]_\times  J \omega_s(t))
\end{bmatrix},
\end{equation*}
where $\omega_s(t)$ and $J$ are the actual angular velocity and the inertia matrix of the MAV, $k_{\omega}$ is the low-level controller gain, while $f(t)$ and $\omega(t)$ are the commanded total thrust and body rate signals.}
Notice that the proposed architecture can be deployed without the need for a dedicated attitude estimator (i.e., using only relative position and gyro measurements), as opposed to other RL-based schemes employing a separate loop for attitude control (see, e.g., \cite{hwangbo2017control}). It is also worth remarking that the validation environment differs from that employed for training the DRL agent.

A first test has been carried out to verify how the controller behaves in an ideal scenario featuring no measurement noise ($w=0$) and nominal model parameters ($\alpha=1$, $\delta=0$). In this scenario, the parameters defining the target trajectory are taken from the same distribution used during training. Figure \ref{nominaltrajectory} displays the trajectory of the tracker for an example simulation. It can be seen that the DRL agent has successfully learned an effective tracking policy and it can follow the target with a suitable heading. Figure \ref{nominaltracking} depicts the evolution of the tracking error and of the distance between the target and the tracker. The former is kept below $0.05$ m at steady-state, while the latter remains always above the minimum allowed value of $y_m=0.4$ m  (i.e., outside the keep-out zone used for collision avoidance). It can be concluded that the controller allows for precise and collision-free tracking operations and that the proposed learning approach provides an effective way to optimize the control performance.

A Monte Carlo simulation approach is adopted to verify the robustness of the DRL policy in off-nominal scenarios featuring measurement noise, model uncertainties, and unseen training data. In particular, we test different combinations of the parameters $\alpha$ and $\delta$, taken from a suitable discretization grid. For each sampled pair of values, we perform 20 runs (lasting 40 seconds), each one featuring a different target trajectory. In order to validate our approach also on target trajectories never experienced during training, we extend the sinusoidal family in \eqref{referencetrajectory} with ramp signals (i.e., linear paths). 

As a comparison baseline, we employ a model-based controller that couples feedback linearization and LQG control. We make this choice due to the lack of fully nonlinear control methods addressing stochastic output feedback from a systematic standpoint within the model-based context. The LQG weights have been tuned extensively to achieve a fair trade-off between performance and robustness. It is worth noticing that the main idea behind the baseline approach is to convert the target tracking problem into a simpler problem in which the relative position and the heading angle are controlled independently (see, e.g., \cite{lotufo2019control}, where a similar approach is pursued). To enable such a design, the absolute orientation of the tracker MAV must be known (in practice we provide it with ground-truth attitude information). Hence, the LQG strategy is favored in the comparison.
\input{figures/real_experiments.tex}

\changes{The metrics employed to compare the performance of the two approaches are the mean and the variance of the tracking error eq.~\eqref{eq:error_track}, averaged over the Monte Carlo runs}. These quantify the tracker’s ability to maintain the desired configuration relative to the target. The results of the comparison are reported in Table \ref{tab:results}. The LQG controller obtains a better score on scenarios close to the nominal one. This is an expected result since the LQG control strategy is optimal for the nominal model and, as mentioned above, it has access to privileged information, i.e.,  the attitude of the tracker MAV. However, it should be noted that the DRL controller deviates from the LQG performance only by a few centimeters, which confirms the effectiveness of the proposed method. On the other hand, the DRL policy obtains better results when one looks at the table \changes{corners} (i.e., for higher model uncertainties). Indeed, under severe off-nominal conditions, the DRL controller shows little performance degradation and it can overcome the model-based counterpart. This is a direct consequence of the robust policy learned. Furthermore, as shown in Figure \ref{perturbedtrajectory}, in the worst-case off-nominal scenario (corresponding to the lower left corner of Table \ref{tab:results}) the LQG controller diverges while the tracking performance of the DRL agent is still acceptable and the collision avoidance constraints are met (see Figure \ref{tracking_worstcase}). 

 \changes{
To investigate the suitability of the proposed method for vision-based target tracking, we rendered the same trajectories reported in Figures \ref{nominaltrajectory} and \ref{perturbedtrajectory}, by using the photo-realistic graphics engine Unreal Engine 4 \cite{unrealengine}. Figure \ref{fig:exp_render} shows some snapshots of the MAVs in the World frame (third-person view), as well as the corresponding images of the target as captured by a virtual camera installed onboard the tracker and aligned with the tracker body-fixed frame (first-person view). It can be seen that the choice of the error function eq.~\eqref{eq:error_track} and the reward term in eqs.~\eqref{rew:track} and~\eqref{rew:axis} allow the tracker to maintain the target within the camera field of view even in the worst-case off-nominal scenario.}

\section{CONCLUSION} \label{conclusions}
In this paper, we presented a model-free control approach based on deep reinforcement learning for target tracking applications involving MAVs. This extends previous results on  robust state feedback to the nonlinear output feedback framework. A Monte Carlo simulation campaign shows that the proposed controller achieves a good tracking performance across a wide variety of operating conditions, outperforming LQG approaches in terms of robustness to uncertainty and noise. \changes{In particular, we found out that the application of domain randomization to the classical gain and delay margin parameters provides an effective way to improve the robustness of the learned policy, without penalizing too much the flight performance.} The obtained results are promising and open up the way for numerous interesting research avenues. 
Future work will focus on rigorously characterizing the stability and robustness properties of the control policy and on validating the controller in real-world settings.

\balance

\bibliographystyle{IEEEtran}
\bibliography{bibliografia}

\end{document}

%% file: figures/overview.tex
\begin{figure}[t]
    \centering
    \includegraphics[width=\linewidth]{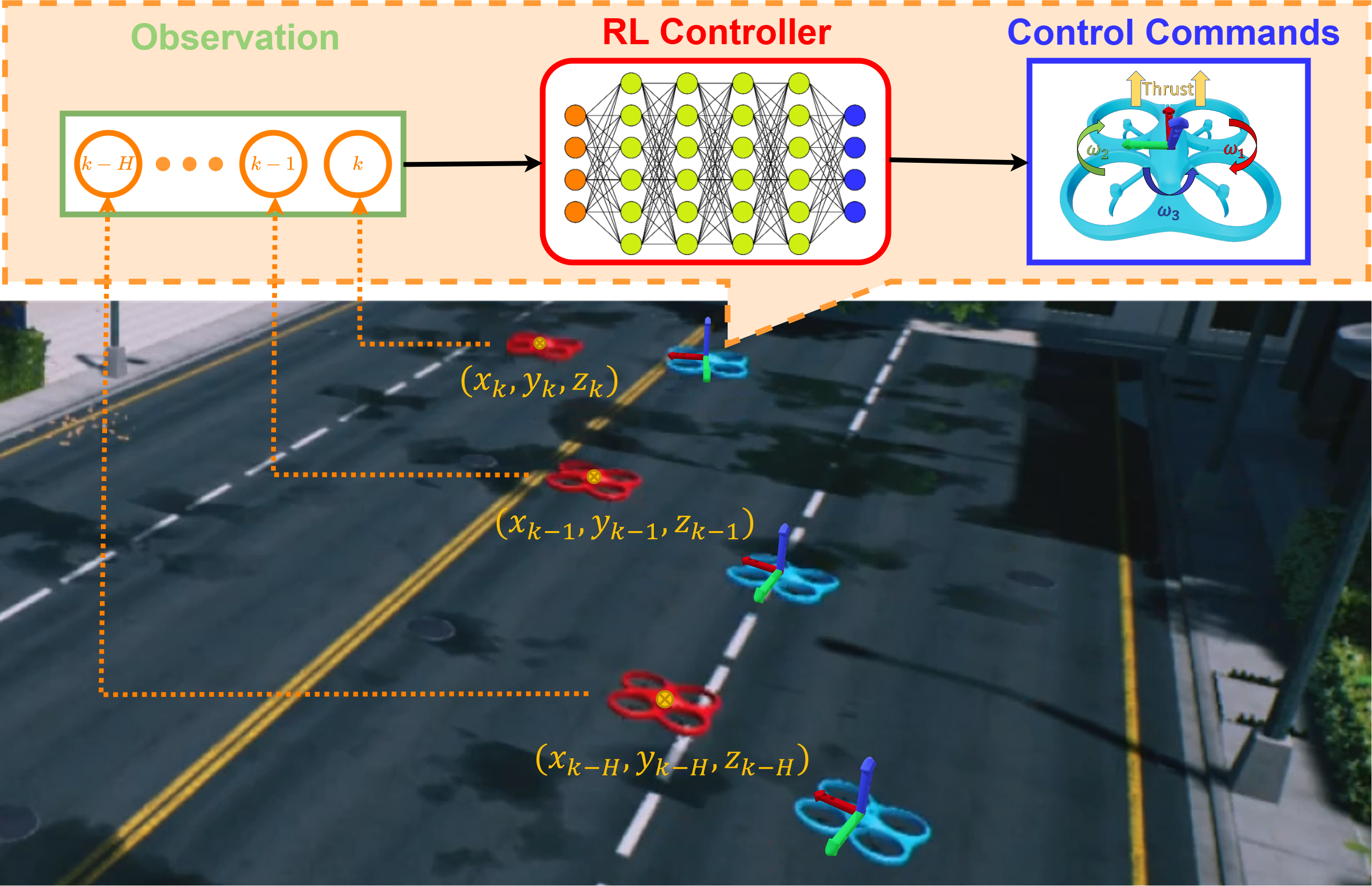}
    \vspace{-1.5em}
    \caption{Target tracking task. The tracker (blue) follows the target (red) while maintaining attitude alignment.}
    \vspace{-1.5em}
    \label{fig:overview}
\end{figure}

%% file: tables/hyper.tex
\begin{table}[t]
\renewcommand{\arraystretch}{1.3}
\caption{Hyperparameters and settings}
\label{table:hyper}
\centering
\begin{tabular}{cc}
\hline
\textbf{Hyperparameter} & \textbf{Value} \\
\hline
Gain parameter $\alpha$ & $\left[0.6, 1.4\right]$ \\
Time delay $\delta$ & $\left[0, 50\right]$ ms\\
Sampling time $t_s$ & $0.05$ s\\
Nominal mass $m_0$ & $1$ kg\\
Reward exponent $\beta$ & $1/3$ \\
Reward coefficients $k_v, k_u$ & $0.4$ \\
Reward constant $c$ & $10$ \\
Desired set-point $y_{r}$ & $\left[75\; 0\; 0 \right]^T$ cm \\
Min. allowed distance $y_{m}$ & $40$ cm \\
Noise std.\ dev. $\sigma_w$ & $0.3$ cm \\
Target spawn std.\ dev. $\sigma_s$ & $10$ cm \\
Target traj. amplitude $A_x, A_y, A_z $ & $\left[1, 30\right]$ m \\
Target traj. frequency $f_x, f_y, f_z$ & $\left[0.002, 0.2\right]$ Hz \\
Target traj. phase $\phi_x, \phi_y, \phi_z$ & $\left[0, 2\pi\right]$ \\
Number of agents & $8$ \\
Learning rate & $0.0003$ \\
Batch size & $256$ \\
Max. episode length & $40$ s \\
Replay buffer size & $1,000,000$ samples \\
Angular velocities  $\omega$ saturation & $\left[-4, 4\right]$ rad/s \\
Thrust increment $\lambda$ saturation & $\left[-20, 20\right]$ N/s \\
Observation sequence length $H$ & $15$ steps\\
Discount factor $\gamma$ & $0.99$ \\
\hline
\end{tabular}
\vspace{-1mm}
\end{table}

%% file: figures/nominal_trajectory.tex
\begin{figure}[t]
    \centering
    \includegraphics[width=0.9\columnwidth]{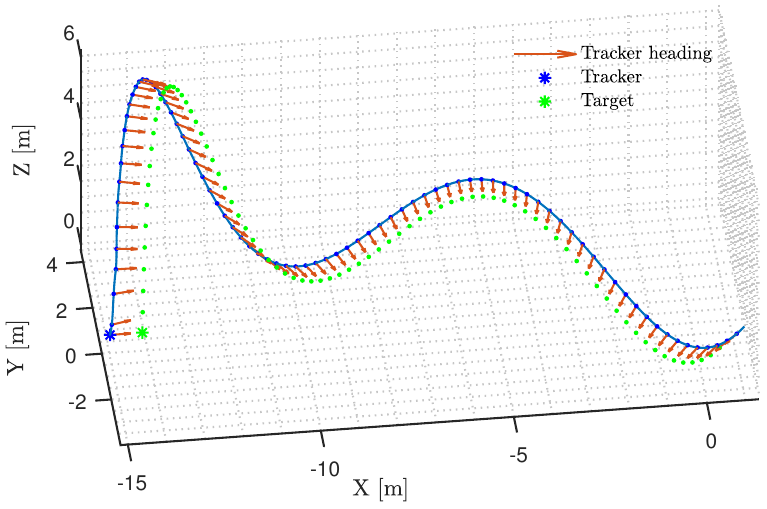}
    \caption{Example of a MAV trajectory obtained by applying the DRL policy in the nominal scenario $\alpha=1$, $\delta=0$ ms.}\label{nominaltrajectory} 
\end{figure}

%% file: figures/tracking_nominal.tex
\begin{figure}[t]
    \centering
    \includegraphics[width=0.9\columnwidth]{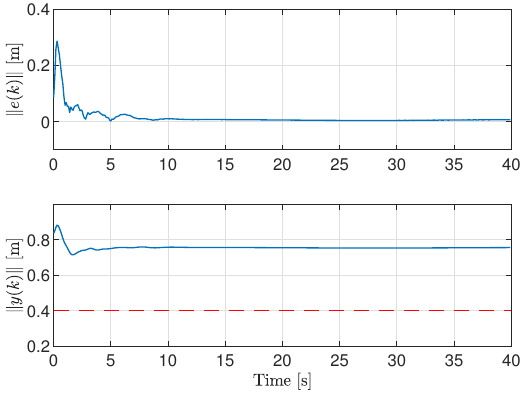}
    \caption{Evolution of the tracking error and of the relative distance obtained by applying the DRL policy in the nominal scenario  $\alpha=1$, $\delta=0$. The keep-out radius is depicted in red (dashed).} \label{nominaltracking} 
    \vspace{-1mm}
\end{figure}

%% file: figures/perturbed_trajectory.tex
\begin{figure}[!t] 
    \centering
    \includegraphics[width=\columnwidth]{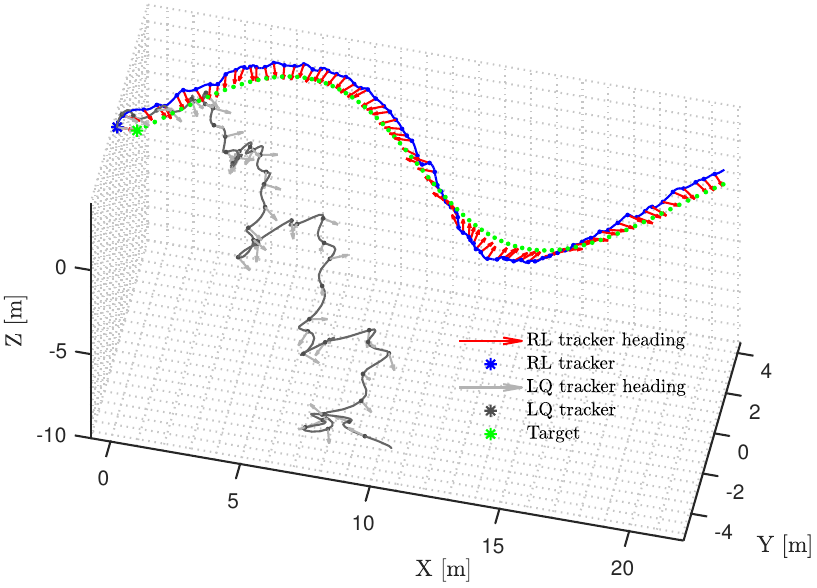}
    \caption{MAV trajectories obtained by applying the DRL and LQG policies in the wort-case scenario $\alpha=0.6$, $\delta=50$ ms.}
    \label{perturbedtrajectory}
\end{figure}

%% file: figures/tracking_worstcase.tex
\begin{figure}[t]
    \centering\vspace{3pt}
    \includegraphics[width=0.9\columnwidth]{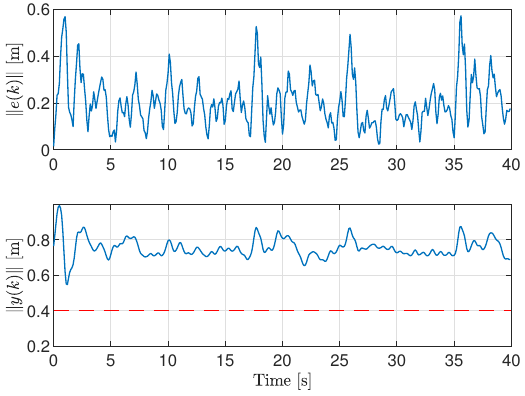}
    \caption{Evolution of the tracking error and of the relative distance obtained by applying the DRL policy in the worst-case scenario $\alpha=0.6$, $\delta=50$ ms. The keep-out radius is depicted in red (dashed).} 
    \label{tracking_worstcase}
    \vspace{-1mm}
\end{figure}

%% file: tables/results.tex
\begin{table*}[!t]
\renewcommand{\arraystretch}{1.3}
\centering
\caption{Experimental results for the mean and the std. dev. of the tracking error}
\resizebox{2\columnwidth}{!}{
\label{tab:results}
\begin{tabular}{c|c|c|c|c|c|c|c|c|c|c|c|c|c|c|c|c|c|c|c}
\cline{3-20}
\multirow{3}{*}{Time Delay $\delta$} & \multirow{3}{*}{Method} & \multicolumn{16}{c}{Gain Parameter $\alpha$}\\
\cline{3-20}
&& \multicolumn{2}{c|}{0.6} & \multicolumn{2}{c|}{0.7} & \multicolumn{2}{c|}{0.8} & \multicolumn{2}{c|}{0.9} & \multicolumn{2}{c|}{1.0} & \multicolumn{2}{c|}{1.1} & \multicolumn{2}{c|}{1.2} & \multicolumn{2}{c|}{1.3} & \multicolumn{2}{c}{1.4}\\
\cline{3-20}
&& $\mu$ & $\sigma$ & $\mu$ & $\sigma$ & $\mu$ & $\sigma$ & $\mu$ & $\sigma$ & $\mu$ & $\sigma$ & $\mu$ & $\sigma$ & $\mu$ & $\sigma$ & $\mu$ & $\sigma$ & $\mu$ & $\sigma$ \\
\hline
\multirow{2}{*}{0 ms} & RL & $ \cellcolor{green!40} \textbf{5.6}$ & $ \cellcolor{green!40} \textbf{8.2}$ & $ \cellcolor{green!40} \textbf{4.1}$ & $ \cellcolor{green!40} \textbf{5.1}$ & $ \cellcolor{red!30} 3.8$ & $ \cellcolor{green!40} \textbf{4.9}$ & $ \cellcolor{red!30} 3.7$ & $ \cellcolor{red!30} 5.0$ & $ \cellcolor{red!30} 3.2$ & $ \cellcolor{red!30} 3.5$ & $ \cellcolor{red!30} 2.9 $ & $ \cellcolor{green!40} \textbf{2.8}$ & $ \cellcolor{green!40} \textbf{3.3}$ & $ \cellcolor{green!40} \textbf{4.1}$ & $ \cellcolor{green!40} \textbf{3.5}$ & $ \cellcolor{green!40} \textbf{5.3}$ & $ \cellcolor{green!40} \textbf{4.7}$ & $ \cellcolor{green!40} \textbf{7.5}$ \\
& LQ & $ \cellcolor{green!40} 5.8$ & $ \cellcolor{green!40} 15.1$ & $ \cellcolor{green!40} 4.2$ & $ \cellcolor{green!40} 10.9$ & $ \cellcolor{red!30} \textbf{2.9}$ & $ \cellcolor{green!40} 7.4$ & $ \cellcolor{red!30} \textbf{1.6}$ & $ \cellcolor{red!30} \textbf{4.2}$ & $ \cellcolor{red!30} \textbf{1.0}$ & $ \cellcolor{red!30} \textbf{2.9}$ & $ \cellcolor{red!30} \textbf{2.1}$ & $ \cellcolor{green!40} 5.5$ & $ \cellcolor{green!40} 3.5$ & $ \cellcolor{green!40} 9.1$ & $ \cellcolor{green!40} 4.9$ & $ \cellcolor{green!40} 13.1$ & $ \cellcolor{green!40} 6.3$ & $ \cellcolor{green!40} 17.2$\\
\hline
\multirow{2}{*}{10 ms} & RL & $ \cellcolor{red!30} 5.8$ & $ \cellcolor{green!40} \textbf{7.4}$ & $ \cellcolor{red!30} 4.2$ & $ \cellcolor{green!40} \textbf{5.2}$ & $ \cellcolor{red!30} 3.7$ & $ \cellcolor{green!40} \textbf{4.6}$ & $ \cellcolor{red!30} 3.3$ & $ \cellcolor{green!40} \textbf{3.7}$ & $ \cellcolor{red!30} 3.1$ & $ \cellcolor{red!30} 3.2$  & $ \cellcolor{red!30} 3.0$ & $ \cellcolor{green!40} \textbf{3.3}$ & $ \cellcolor{green!40} \textbf{3.3}$ & $ \cellcolor{green!40} \textbf{4.4}$ & $ \cellcolor{green!40} \textbf{3.7}$ & $ \cellcolor{green!40} \textbf{5.9}$ & $ \cellcolor{green!40} \textbf{4.7}$ & $ \cellcolor{green!40} \textbf{7.9}$\\
 & LQ & $ \cellcolor{red!30} \textbf{5.3}$ & $ \cellcolor{green!40} 14.2$ & $ \cellcolor{red!30} \textbf{3.9}$ & $ \cellcolor{green!40} 10.3$ & $ \cellcolor{red!30} \textbf{2.7}$ & $ \cellcolor{green!40} 7.0$ & $ \cellcolor{red!30} \textbf{1.5}$ & $ \cellcolor{green!40} 4.0$ & $ \cellcolor{red!30} \textbf{1.1}$ & $ \cellcolor{red!30} \textbf{3.1}$  & $ \cellcolor{red!30} \textbf{2.2}$ & $ \cellcolor{green!40} 5.8$ & $ \cellcolor{green!40} 3.6$ & $ \cellcolor{green!40} 9.4$ & $ \cellcolor{green!40} 5.0$ & $ \cellcolor{green!40} 13.4$ & $ \cellcolor{green!40} 6.4$ & $ \cellcolor{green!40} 17.5$\\
\hline
\multirow{2}{*}{20 ms} & RL & $ \cellcolor{red!30} 6.5$ & $ \cellcolor{green!40} \textbf{7.1}$ & $ \cellcolor{red!30} 5.5$ & $ \cellcolor{green!40} \textbf{5.6}$ & $ \cellcolor{red!30} 4.4$ & $ \cellcolor{green!40} \textbf{5.3}$ & $ \cellcolor{red!30} 3.5$ & $ \cellcolor{red!30} 4.3$ & $ \cellcolor{red!30} 3.2$ & $ \cellcolor{red!30} 3.8$  & $ \cellcolor{red!30} 3.3$ & $ \cellcolor{green!40} \textbf{4.0}$ & $ \cellcolor{green!40} \textbf{3.4}$ & $ \cellcolor{green!40} \textbf{5.0}$ & $ \cellcolor{green!40} \textbf{4.1}$ & $ \cellcolor{green!40} \textbf{6.6}$ & $ \cellcolor{green!40} \textbf{5.0}$ & $ \cellcolor{green!40} \textbf{8.7}$\\
 & LQ & $ \cellcolor{red!30} \textbf{5.0}$ & $ \cellcolor{green!40} 13.4$ & $ \cellcolor{red!30} \textbf{3.5}$ & $ \cellcolor{green!40} 9.8$ & $ \cellcolor{red!30} \textbf{2.4}$ & $ \cellcolor{green!40} 6.7$ & $ \cellcolor{red!30} \textbf{1.4}$ & $ \cellcolor{red!30} \textbf{3.9}$ & $ \cellcolor{red!30} \textbf{1.2}$ & $ \cellcolor{red!30} \textbf{3.4}$  & $ \cellcolor{red!30} \textbf{2.4}$ & $ \cellcolor{green!40} 6.1$ & $ \cellcolor{green!40} 3.7$ & $ \cellcolor{green!40} 9.8$ & $ \cellcolor{green!40} 5.1$ & $ \cellcolor{green!40} 13.7$ & $ \cellcolor{green!40} 6.5$ & $ \cellcolor{green!40} 17.8$\\
\hline
\multirow{2}{*}{30 ms} & RL & $ \cellcolor{red!30} 11.2$ & $ \cellcolor{green!40} \textbf{8.0}$ & $ \cellcolor{red!30} 6.6$ & $ \cellcolor{green!40} \textbf{6.7}$ & $ \cellcolor{red!30} 5.8$ & $ \cellcolor{red!30} 6.5$ & $ \cellcolor{red!30} 4.8$ & $ \cellcolor{red!30} 6.6$ & $ \cellcolor{red!30} 3.9$ & $ \cellcolor{red!30} 5.2$  & $ \cellcolor{red!30} 3.9$ & $ \cellcolor{green!40} \textbf{5.5}$ & $ \cellcolor{red!30} 3.9$ & $ \cellcolor{green!40} \textbf{6.1}$ & $ \cellcolor{green!40} \textbf{4.6}$ & $ \cellcolor{green!40} \textbf{7.8}$ & $ \cellcolor{green!40} \textbf{5.0}$ & $ \cellcolor{green!40} \textbf{9.1}$ \\
 & LQ & $ \cellcolor{red!30} \textbf{7.3}$ & $ \cellcolor{green!40} 14.7$ & $ \cellcolor{red!30} \textbf{3.3}$ & $ \cellcolor{green!40} 9.3$ & $ \cellcolor{red!30} \textbf{2.3}$ & $ \cellcolor{red!30} \textbf{6.3}$ & $ \cellcolor{red!30} \textbf{1.3}$ & $ \cellcolor{red!30} \textbf{3.9}$ & $ \cellcolor{red!30} \textbf{1.4}$ & $ \cellcolor{red!30} \textbf{4.0}$  & $ \cellcolor{red!30} \textbf{2.5}$ & $ \cellcolor{green!40} 6.9$ & $ \cellcolor{red!30} \textbf{3.8}$ & $ \cellcolor{green!40} 10.5$ & $ \cellcolor{green!40} 5.2$ & $ \cellcolor{green!40} 14.4$ & $ \cellcolor{green!40} 6.6$ & $ \cellcolor{green!40} 18.4$\\
\hline
\multirow{2}{*}{40 ms} & RL & $ \cellcolor{green!40} \textbf{17.7}$ & $ \cellcolor{green!40} \textbf{10.8}$ & $ \cellcolor{red!30} 11.6$ & $ \cellcolor{green!40} \textbf{8.2}$ & $ \cellcolor{red!30} 7.7$ & $ \cellcolor{red!30} 8.6$ & $ \cellcolor{red!30} 6.4$ & $ \cellcolor{red!30} 7.0$ & $ \cellcolor{red!30} 5.0$ & $ \cellcolor{red!30} 6.4$  & $ \cellcolor{red!30} 4.8$ & $ \cellcolor{green!40} \textbf{6.3}$ & $ \cellcolor{red!30} 4.8$ & $ \cellcolor{green!40} \textbf{7.1}$ & $ \cellcolor{green!40} \textbf{5.2}$ & $ \cellcolor{green!40} \textbf{8.6}$ & $ \cellcolor{green!40} \textbf{5.4}$ & $ \cellcolor{green!40} \textbf{9.9}$ \\
& LQ & $ \cellcolor{green!40} 54.9$ & $ \cellcolor{green!40} 26.7$ & $ \cellcolor{red!30} \textbf{3.4}$ & $ \cellcolor{green!40} 9.2$ & $ \cellcolor{red!30} \textbf{2.2}$ & $ \cellcolor{red!30} \textbf{6.2}$ & $ \cellcolor{red!30} \textbf{1.3}$ & $ \cellcolor{red!30} \textbf{3.9}$ & $ \cellcolor{red!30} \textbf{1.5}$ & $ \cellcolor{red!30} \textbf{4.0}$  & $ \cellcolor{red!30} \textbf{2.7}$ & $ \cellcolor{green!40} 6.9$ & $ \cellcolor{red!30} \textbf{4.0}$ & $ \cellcolor{green!40} 10.5$ & $ \cellcolor{green!40} 5.3$ & $ \cellcolor{green!40} 14.4$ & $ \cellcolor{green!40} 6.7$ & $ \cellcolor{green!40} 18.4$\\
\hline
\multirow{2}{*}{50 ms} & RL & $ \cellcolor{green!40} \textbf{22.8}$ & $ \cellcolor{green!40} \textbf{12.4}$ & $ \cellcolor{red!30} 15.1$ & $ \cellcolor{green!40} \textbf{9.5}$ & $ \cellcolor{red!30} 11.8$ & $ \cellcolor{red!30} 9.4$ & $ \cellcolor{red!30} 7.6$ & $ \cellcolor{red!30} 7.6$ & $ \cellcolor{red!30} 6.6$ & $ \cellcolor{red!30} 6.5$  & $ \cellcolor{red!30} 5.8$ & $ \cellcolor{red!30} 7.4$ & $ \cellcolor{red!30} 6.4$ & $ \cellcolor{green!40} \textbf{9.3}$ & $ \cellcolor{red!30} 5.8$ & $ \cellcolor{green!40} \textbf{9.4}$ & $ \cellcolor{green!40} \textbf{6.1}$ & $ \cellcolor{green!40} \textbf{11.1}$\\
& LQ & $ \cellcolor{green!40} 860.1$ & $ \cellcolor{green!40} 470.5$ & $ \cellcolor{red!30} \textbf{4.5}$ & $ \cellcolor{green!40} 10.3$ & $ \cellcolor{red!30} \textbf{2.3}$ & $ \cellcolor{red!30} \textbf{6.3}$ & $ \cellcolor{red!30} \textbf{1.5}$ & $ \cellcolor{red!30} \textbf{4.2}$ & $ \cellcolor{red!30} \textbf{1.8}$ & $ \cellcolor{red!30} \textbf{4.5}$  & $ \cellcolor{red!30} \textbf{2.9}$ & $ \cellcolor{red!30} \textbf{7.3}$ & $ \cellcolor{red!30} \textbf{4.2}$ & $ \cellcolor{green!40} 10.9$ & $ \cellcolor{red!30} \textbf{5.5}$ & $ \cellcolor{green!40} 14.8$ & $ \cellcolor{green!40} 6.8$ & $ \cellcolor{green!40} 18.8$\\
\hline
\end{tabular}}
 \vspace{-1mm}
\end{table*}

%% file: figures/real_experiments.tex
\begin{figure*}[t]
    \centering
    \subfigure[]{
    \includegraphics[width=0.48\textwidth]{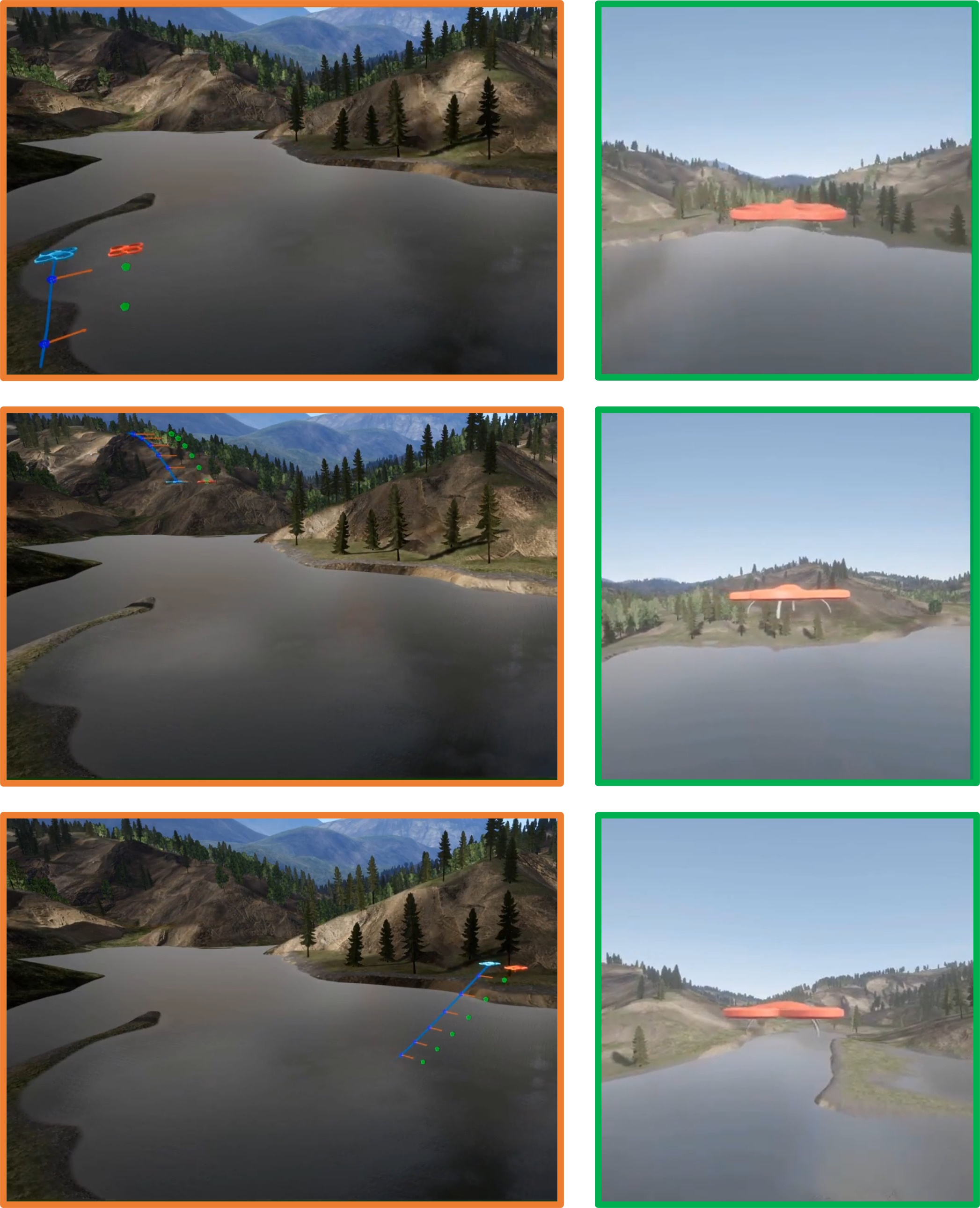}
    \label{fig:nominal_exp}
    }
    \hfill
    \subfigure[]{
    \includegraphics[width=0.48\textwidth]{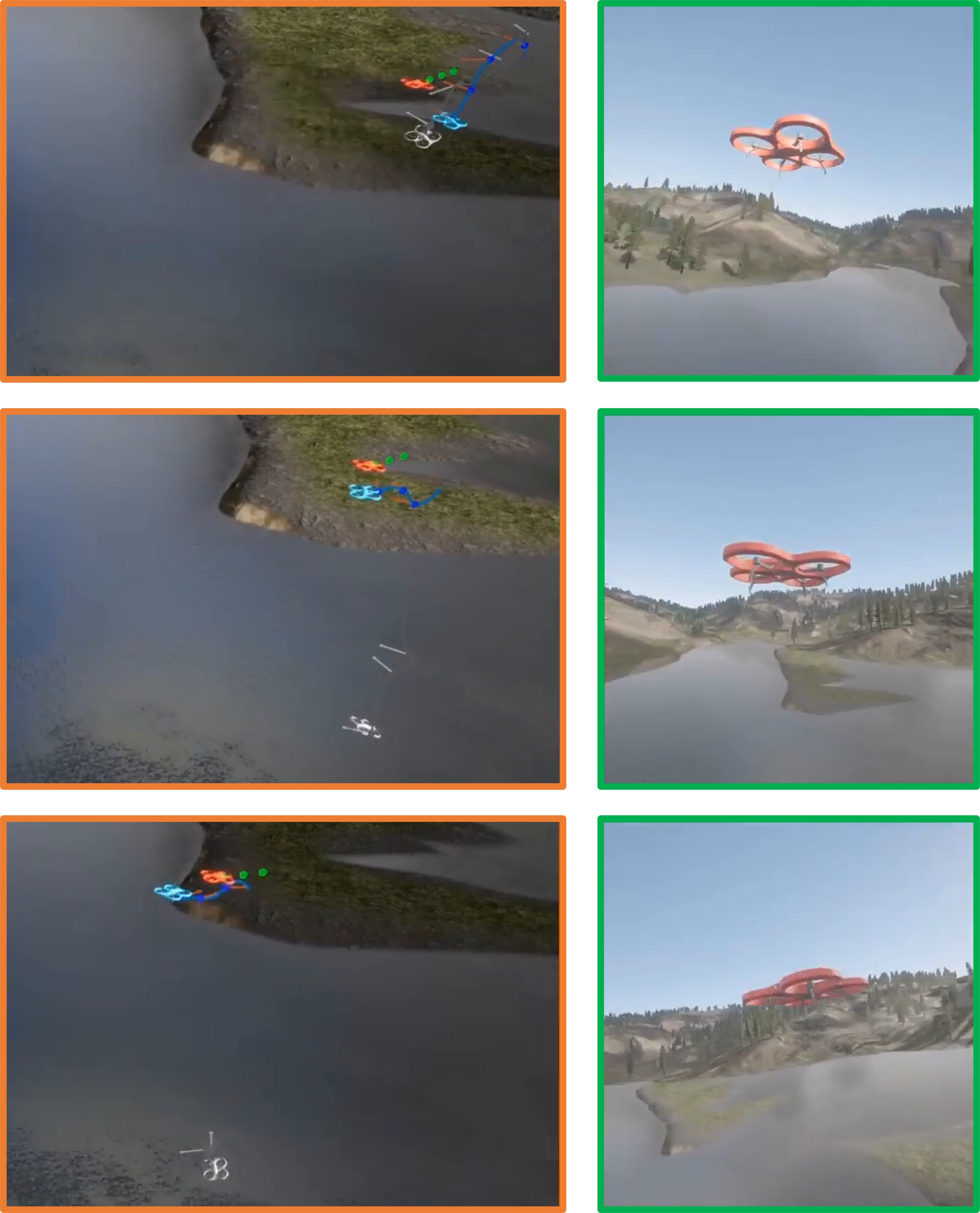}
    \label{fig:noise_exp}
    }
    \caption{Rendering of the two trajectories depicted in Figures \ref{nominaltrajectory} and \ref{perturbedtrajectory}: (a) nominal scenario $\alpha=1$, $\delta=0$. (b) worst-case off-nominal scenario $\alpha=0.6$, $\delta=50$ ms. In both cases, the tracker MAV equipped with the DRL controller is colored in blue, while the target is colored in red. The third-person view is edged in orange and the first-person view of the blue tracker is edged in green. In the third-person view corresponding to the off-nominal scenario, the tracker MAV equipped with the baseline LQG controller is also depicted (in gray).}
    \label{fig:exp_render}\vspace{-1mm}
\end{figure*}